\documentclass[10pt,twocolumn,letterpaper]{article}

\usepackage[pagenumbers]{wacv} 

\usepackage{graphicx}
\usepackage{amsmath}
\usepackage{amssymb}
\usepackage{booktabs}
\usepackage{multirow}

\usepackage{ragged2e}
\usepackage{bm}

\usepackage{amsfonts}
\usepackage{algorithm}
\usepackage{algpseudocode}

\usepackage{diagbox}
\usepackage{comment}

\usepackage[accsupp]{axessibility} 

\definecolor{wacvblue}{rgb}{0.21,0.49,0.74}
\usepackage[pagebackref,breaklinks,colorlinks,allcolors=wacvblue]{hyperref}

\def\wacvPaperID{440} 
\def\confName{WACV}
\def\confYear{2026}

\title{GFT-GCN: Privacy-Preserving 3D Face Mesh Recognition with Spectral Diffusion}

\author{Hichem Felouat$^{1,2}$ \quad\quad Hanrui Wang$^{2}$ \quad\quad Isao Echizen$^{1,2,3}$\\
$^{1}$The Graduate University for Advanced Studies, SOKENDAI, Kanagawa, Japan\\
$^{2}$National Institute of Informatics, NII, Tokyo, Japan\\
$^{3}$The University of Tokyo, Tokyo, Japan\\
{\tt\small \{hichemfel, hanrui\_wang, iechizen\}@nii.ac.jp}
}

\begin{document}
\maketitle

\begin{abstract}
3D face recognition offers a robust biometric solution by capturing facial geometry, providing resilience to variations in illumination, pose changes, and presentation attacks. Its strong spoof resistance makes it suitable for high-security applications, but protecting stored biometric templates remains critical. We present GFT-GCN, a privacy-preserving 3D face recognition framework that combines spectral graph learning with diffusion-based template protection. Our approach integrates the Graph Fourier Transform (GFT) and Graph Convolutional Networks (GCN) to extract compact, discriminative spectral features from 3D face meshes. To secure these features, we introduce a spectral diffusion mechanism that produces irreversible, renewable, and unlinkable templates. A lightweight client-server architecture ensures that raw biometric data never leaves the client device. Experiments on the BU-3DFE and FaceScape datasets demonstrate high recognition accuracy and strong resistance to reconstruction attacks. Results show that GFT-GCN effectively balances privacy and performance, offering a practical solution for secure 3D face authentication. The code is available at \url{https://github.com/hichemfelouat/GFT-GCN}.

\end{abstract}
\section{Introduction}
\label{sec:intro}

\begin{figure}[t]
     \center
     \includegraphics[scale=0.23]{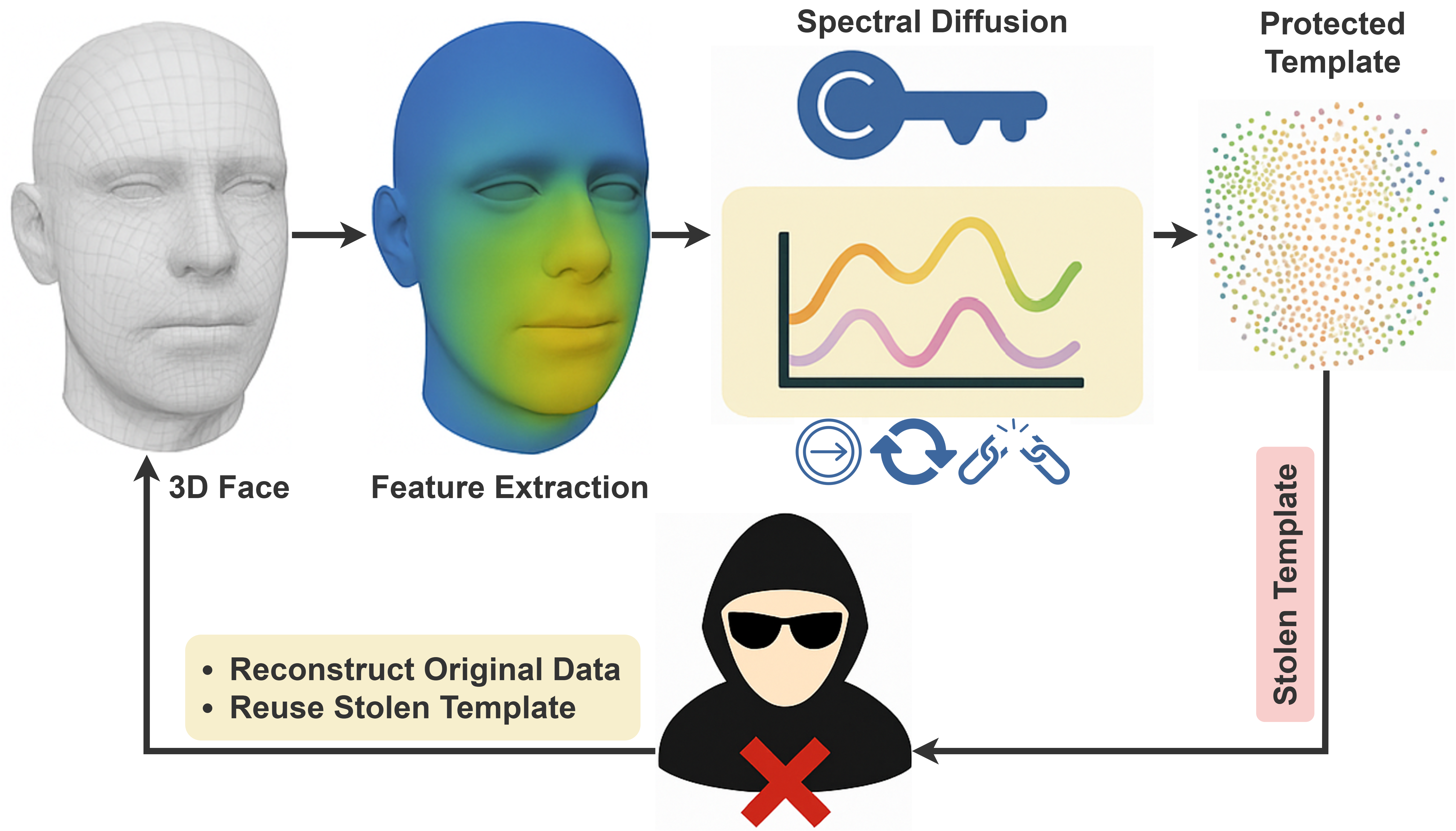}
     \caption{The GFT-GCN framework secures spectral features from 3D face meshes using spectral diffusion, generating templates that are irreversible, renewable, and unlinkable. Protected templates cannot reveal the original biometric data or be reused while maintaining high recognition accuracy for genuine users.}
     \label{fig:abstract_graphical}
\end{figure}

Biometric systems are increasingly used for secure and convenient identity verification in various applications, including border control, banking, and personal devices. Among different biometric modalities, 3D face recognition has gained attention as a reliable alternative to 2D methods \cite{guo20233d,li2022comprehensive, jing20233d, jing20213d, gilani2018learning}. Unlike 2D images, 3D face data captures geometric information such as depth, shape, and surface curvature \cite{papadopoulos2022face, dong2023laplacian2mesh, dong2024task}. This makes 3D face recognition more robust to lighting changes, pose variations, and facial texture differences  \cite{yu2025adaptive, xu2024depth, yang2024pointsurface, jing20233d}. In addition to improving recognition accuracy, 3D face data also strengthens resistance to presentation attacks, including printed photos, video replays, and even forged 3D masks \cite{felouat20253ddgd, dommati2023real, kong2025not, yang2023towards}. However, as the use of biometric systems expands, concerns about user privacy and data security are growing. Unlike passwords, biometric traits are permanent; they cannot be changed or revoked if compromised \cite{abdullahi2024biometric, zhou2007template}. For example, a person may rely on 3D face recognition for border control, banking, and access to facilities. Using the same biometric trait across multiple systems increases risk: if a single template is leaked, identity fraud may occur across all connected services. Therefore, protecting biometric templates is essential to ensure both privacy and system reliability.

Biometric Template Protection (BTP) schemes aim to address this challenge. Effective BTP methods must ensure (1) high recognition accuracy, (2) irreversibility, preventing reconstruction of the original data, and (3) renewability and unlinkability, enabling re-issuance of independent templates without cross-system traceability \cite{abdullahi2024biometric, hahn2022biometric}.  Most existing BTP methods are designed for 2D features. When applied to 3D face data, they often fail to accommodate the geometric nature of 3D faces, resulting in limited irreversibility, inadequate support for renewability across applications, and increased vulnerability to variability and noise \cite{abdullahi2024biometric, hahn2022biometric}.

In this paper, we present GFT-GCN, a privacy-preserving framework for 3D face mesh recognition (Figure~\ref{fig:abstract_graphical}). Our method begins by extracting spectral features using the Graph Fourier Transform (GFT), which captures geometric structure in a compact and efficient manner. These features are then refined using a Graph Convolutional Network (GCN) to enhance discriminability for recognition. To ensure privacy, we introduce a spectral diffusion process that transforms features into protected templates that are irreversible, renewable, and unlinkable. The framework follows a lightweight client-server design, where only protected templates are transmitted, preserving user privacy while maintaining strong recognition performance.

Our main contributions are: (1) We propose a novel integration of GFT and GCN for efficient spectral feature learning on 3D face meshes, improving recognition accuracy and computational efficiency; (2) We introduce a spectral diffusion mechanism that enforces irreversibility, renewability, and unlinkability, offering a new diffusion-based approach to biometric template protection; (3) We design a privacy-aware client-server architecture that performs all sensitive computations locally; (4)  We conduct extensive experiments on BU-3DFE and FaceScape showing high accuracy and strong resistance to reconstruction attacks; and (5) Finally, we provide a mathematically grounded diffusion process that supports stable and practical deployment. Together, these contributions advance secure 3D face recognition without compromising utility.

\section{Related Work}
\subsection{Biometric Template Protection}
Biometric Template Protection (BTP) is a fundamental aspect of the deployment of secure and privacy-preserving biometric recognition systems. Unlike passwords or tokens, biometric traits are inherently irreplaceable and cannot be revoked if compromised \cite{abdullahi2024biometric, hahn2022biometric}. Thus, any BTP mechanism must meet three critical requirements: (1) high recognition, (2) irreversibility, and (3) renewability and unlinkability.

According to Abdullahi et al. \cite{abdullahi2024biometric}, BTP techniques are generally classified into three categories:

Cancelable Biometrics apply repeatable, non-invertible transformations such as biometric salting or irreversible mapping \cite{bernal2023review}. These methods are efficient and support renewability, but must balance protection strength with feature discriminability.

Biometric Cryptosystems include key binding and key generation approaches that link cryptographic keys to biometric data \cite{kaur2023biometric, sharma2023survey}. Although theoretically secure, they often face challenges with biometric variability and require error correction.

Encrypted Domain Matching enables privacy-preserving computation using homomorphic encryption or garbled circuits \cite{osorio2021stable}. These offer strong protection, but are typically too computationally intensive for real-time biometric authentication.

Our GFT-GCN framework falls within the cancelable biometrics domain, introducing a spectral diffusion mechanism that ensures irreversibility, renewability, and unlinkability while preserving recognition accuracy through advanced graph spectral learning.

\subsection{Graph Fourier Transform}
The Graph Fourier Transform (GFT) extends classical Fourier analysis to graph-structured data, enabling the spectral analysis of signals in non-Euclidean domains \cite{sardellitti2017graph, domingos2020graph}. For a graph \( G = (V, E) \) with vertices \( V \) and edges \( E \), the GFT uses the graph Laplacian, derived from the adjacency matrix \( A \) and degree matrix \( D \), defined as \( L = D - A \). The Laplacian \( L \) is diagonalized into eigenvectors \( U \) and eigenvalues \( \Lambda \), forming the spectral basis. For a signal \( f \in \mathbb{R}^{|V|} \) on the vertices, the GFT is \( F = U^T f \), where \( F \) represents spectral coefficients, low frequencies indicate smooth variations, and high frequencies show rapid changes. The inverse GFT reconstructs the signal as \( f = U F \). GFT is valuable for filtering, denoising, and feature extraction, capturing graph structure effectively. Focusing on low-frequency coefficients enables reduction in dimensionality while retaining key patterns, benefiting areas such as social network analysis and machine learning on graphs \cite{ricaud2019fourier}.

\subsection{Diffusion Models}
Diffusion models are a robust framework for generative tasks, defined by a forward process that adds noise to data and a reverse process that denoises to recover the original distribution. The general equation for the forward diffusion process at step $t$ is:

\begin{equation}
    q(x_t | x_{t-1}) = \mathcal{N}(x_t; \sqrt{1 - \beta_t} x_{t-1}, \beta_t I),
\end{equation}

where $\beta_t$ is the noise schedule, $x_0$ is the original data, and $t = 1, \ldots, T$.

Ho et al. \cite{ho2020denoising} introduced Denoising Diffusion Probabilistic Models (DDPMs), achieving high-fidelity synthesis via a variational bound. Nichol et al. \cite{nichol2021improved} refined noise schedules for improved efficiency. In 3D, Friedrich et al. \cite{friedrich2024wdm} proposed WDM, which performs denoising in wavelet space to preserve spatial structure in medical volumes.

\paragraph{Spectral Diffusion Processes:}
Classical diffusion operates in Euclidean space, whereas spectral diffusion extends it to graphs and manifolds by working in the frequency domain. Phillips et al. \cite{phillips2022spectral} introduced operators that emphasize low-frequency components while preserving topology. Luo et al. \cite{luo2023fast} further developed scalable spectral diffusion for graph generation, demonstrating that spectral diffusion is particularly well-suited for 3D meshes and graph biometrics, where privacy-preserving transformations must retain both identity and topology.

\section{Methodology}
\begin{figure*}[ht]
     \center
     \includegraphics[scale=0.45]{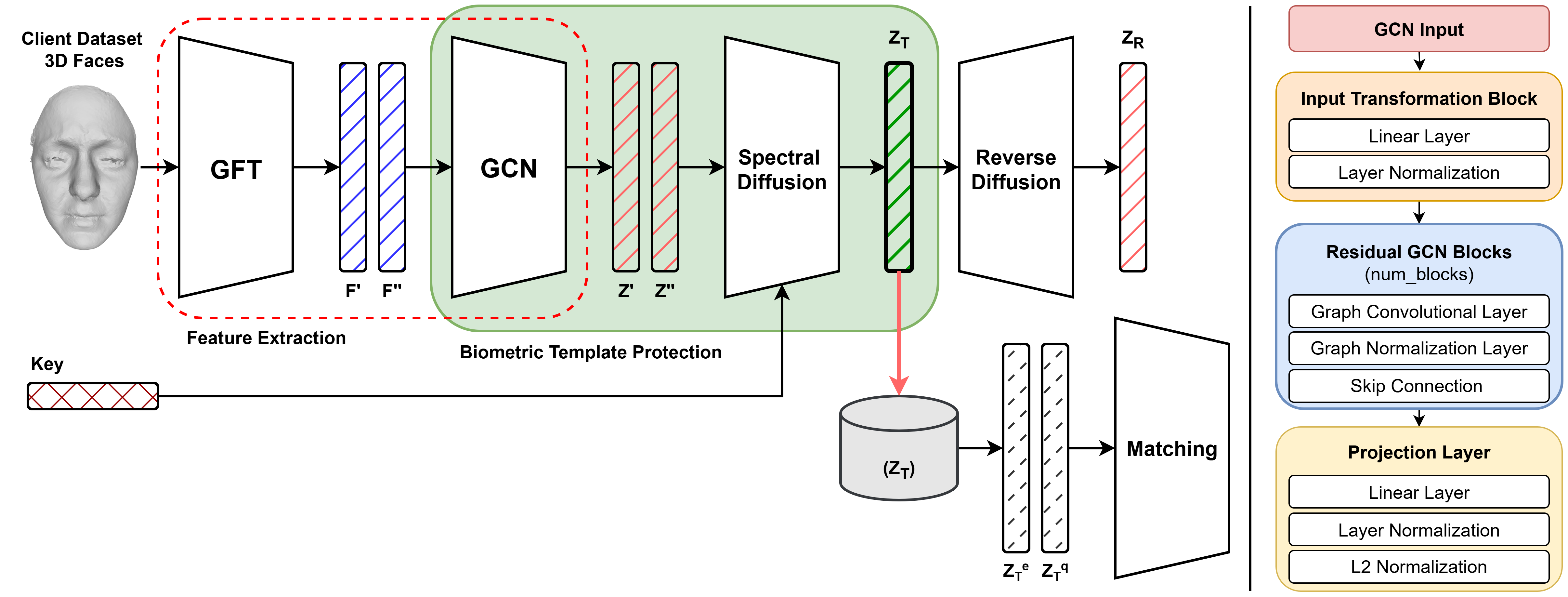}
     \caption{Overview of the GFT-GCN framework for privacy-preserving 3D face recognition. Spectral features are extracted from 3D face meshes using the Graph Fourier Transform (GFT) and refined with a Graph Convolutional Network (GCN). A key-dependent spectral diffusion process transforms the embeddings into protected templates $Z_T$, used for matching between a query $Z_T^q$ and an enrolled $Z_T^e$. Reverse diffusion is applied only during training to guide the model and is not used in inference. The right panel shows the internal structure of the GCN module.}
     \label{fig:GFT_GCN_approach}
\end{figure*}

We propose a privacy-preserving 3D face mesh recognition framework that combines spectral feature extraction, geometric refinement, and template protection within a lightweight client-server architecture (see Figure \ref{fig:GFT_GCN_approach}). Each client trains locally with limited data and computing resources, using methods designed for such constraints. The training consists of two stages: first, a Graph Convolutional Network (GCN) refines the spectral features extracted via the Graph Fourier Transform (GFT); second, a spectral diffusion process generates protected templates using the frozen GCN. The framework ensures privacy through three layers: (1) low-frequency spectral filtering, (2) GCN-based transformation, and (3) spectral diffusion for irreversibility, renewability, and unlinkability. Together, these components enable high recognition accuracy while protecting against reconstruction attacks.

\subsection{GFT-Based Spectral Feature Extraction}
\label{subsec:gft}
The Graph Fourier Transform (GFT) leverages the spectral properties of 3D meshes to extract compact, low-frequency features that capture dominant geometric structures, offering resilience to noise and minor deformations. Given a 3D face mesh \( M = (V, E) \), where \( V \) denotes the set of vertices and \( E \) the edges, we model the mesh as an undirected graph. The extraction process proceeds as follows:

\begin{enumerate}
    \item Normalize Laplacian: \( \tilde{L} = D^{-1/2} L D^{-1/2} \)
    
    \item Perform eigen-decomposition of \( \tilde{L} = U \Lambda U^T \), where \( U \) contains the eigenvectors and \( \Lambda \) the eigenvalues.
    
    \item Extract vertex features \( f \in \mathbb{R}^{|V| \times n} \) (e.g., 3D coordinates \( \{x_i, y_i, z_i\} \)) and compute the GFT: $F = U^T f $
  
    \item Select the low-frequency coefficients \( F_{\text{low}} \in \mathbb{R}^{ k\times n} \), corresponding to the first \( k\) smallest components, to form a reduced representation of the mesh.
\end{enumerate}

This spectral truncation preserves essential shape information while minimizing high-frequency noise, forming the first layer of privacy by limiting the information available for reconstruction.

\subsection{GCN-Based Feature Refinement}
\label{subsec:gcn}
Graph Convolutional Networks (GCNs) excel at modeling geometric relationships in irregular graph structures, making them ideal for refining the spectral features into discriminative biometric descriptors. We employ a GCN to transform \( F_{\text{low}} \) into a feature vector suitable for recognition. The final GCN output, denoted as $Z$, forms the biometric feature vector: $Z = H^{(L)}$, where $L$ is the number of GCN layers. This transformation enhances discriminability while preserving the underlying structure of the face mesh. The GCN is trained end-to-end on the client-labeled dataset of 3D face meshes, optimizing a siamese contrastive loss. Once trained, its weights are frozen, providing a non-linear transformation as the second layer of privacy protection.

\subsubsection{Loss Function for GCN Training}
\label{subsubsec:loss}
The Graph Convolutional Network (GCN) is trained using a Siamese contrastive loss, denoted as $ L_{GCN} $, to enhance the discriminability of the biometric feature vector $Z$. This loss function is designed to optimize the embedding space by encouraging embeddings of the same subject to be closer and embeddings of different subjects to be farther apart, thereby improving recognition accuracy. Given a training set of $N$ labeled pairs, the loss is defined as:

\begin{multline}
\mathcal{L}_{\text{GCN}} = \frac{1}{N} \sum_{k=1}^{N} \big[ 
y_k \left( 1 - d(Z_{k'}, Z_{k''}) \right) \\
+ (1 - y_k) \max \left( 0, \, m - \left(1 - d(Z_{k'}, Z_{k''}) \right) \right) \big]
\end{multline}

where $Z_{k'}, Z_{k''} \in \mathbb{R}^d$ denote the feature vectors of the $k$-th pair, $d(\cdot, \cdot)$ is a distance function based on similarity (e.g., cosine distance), $y_k \in \{0,1\}$ is the binary label that indicates whether the pair belongs to the same identity ($y_k = 1$) or not ($y_k = 0$), and $m > 0$ is a predefined margin that specifies the minimum separation required for negative pairs.

This formulation encourages the following behavior:
\begin{itemize}
    \item For match pairs (\( y_k = 1 \)), the term \( (1 - d(Z_{k'}, Z_{k''})) \) is minimized, encouraging a small distance between embeddings of the same subject, thus enhancing intra-class compactness.
    
    \item For mismatch pairs (\( y_k = 0 \)), the term \( \max(0, m - (1 - d(Z_{k'}, Z_{k''}))) \) pushes the distance beyond the margin $m$, ensuring inter-class separation.
\end{itemize}

The goal of this loss function is to maximize the separation between different subjects and reduce intra-subject variance. The contrastive loss directly enhances the class separability of the learned features. This leads to improved recognition performance, especially under variations such as pose, expression, and occlusion. As a result, the GCN provides a robust, non-linear transformation that serves as an additional layer of privacy protection while maintaining high discriminative power.

\subsection{Spectral Diffusion for BTP}
\label{subsec:diffusion}
We introduce spectral diffusion as a parameterized forward process to protect a biometric feature vector $Z$. This mechanism forms the third layer of privacy. Unlike traditional diffusion models that rely solely on random noise, our method employs a key-conditioned neural transformation $\phi(., k)$ that guides the transformation. This design ensures that the resulting template is both irreversible and renewable: irreversible because the original feature cannot be reconstructed without the user-specific key $k$, and renewable because different keys yield unique, uncorrelated templates for the same subject (see Algorithm~\ref{alg:spectral_diffusion}). For more information, refer to Appendix A. The forward diffusion process is defined as
\begin{equation}
    Z_t = Z_{t-1} + \sqrt{\beta_t}\,\epsilon + (1 - \beta_t)\,\phi(Z_{t-1}, k)
\end{equation}
where $\epsilon \sim \mathcal{N}(0,I)$ is Gaussian noise and $\beta_t$ follows a linear schedule. The network $\phi$, conditioned on $k$, is trained to predict the noise component.  

This structured forward diffusion provides a consistent, key-dependent transformation that cannot be replicated by arbitrary stochastic perturbations. It enables the generation of stable templates for a given key while preserving discriminability, and it guarantees security since recovering $Z$ from $Z_T$ would require solving a multi-step inverse problem under high-dimensional uncertainty. The final protected template $Z_T$ is stored and used for authentication.

\begin{algorithm}[ht]
    \caption{Spectral Diffusion for Template Protection}
    \label{alg:spectral_diffusion}
    \begin{algorithmic}[1]
        \Require Biometric feature vector $Z \in \mathbb{R}^d$, user-specific key $k$, noise schedule $\{\beta_t\}_{t=1}^T$, number of diffusion steps $T$, transformation function $\phi$
        \Ensure Protected template $Z_T$
        \State Initialize $Z_0 \gets Z$ \Comment{Set initial feature vector}
        \For{$t = 1$ to $T$}
            \State Sample noise $\epsilon \sim \mathcal{N}(0, I)$ \Comment{Gaussian noise}
            \State Compute transformation $\phi(Z_{t-1}, k)$ \Comment{Neural network with key $k$}
            \State Update feature vector:
            \[
            Z_t \gets \sqrt{1 - \beta_t} Z_{t-1} + \sqrt{\beta_t} \epsilon + (1 - \beta_t) \phi(Z_{t-1}, k)
            \]
        \EndFor
        \State \Return $Z_T$ \Comment{Protected biometric template}
    \end{algorithmic}
\end{algorithm}

\subsubsection{Loss Functions for Spectral Diffusion Training}
\label{subsubsec:losses}
The spectral diffusion model is optimized with a composite loss framework that balances recognition accuracy and privacy preservation. The primary components are discriminability and contrastive losses, augmented by unlinkability and key diversity losses to ensure template unlinkability and key-dependent diversity.

\paragraph{Discriminability Loss: } This loss preserves the identity-based similarity structure between original embeddings $Z$ and protected templates $Z_T$, while separating impostor pairs. It comprises three terms: preserving similarity for same-subject same-key pairs, suppressing similarity for different-subject same-key pairs, and disassociating same-subject different-key pairs. Let $S_{ij}^{\text{orig}} = \cos(Z_i, Z_j)$ and $S_{ij}^{\text{prot}} = \cos(Z_{T,i}, Z_{T,j})$ denote cosine similarities, with binary masks $M_{ij}^{\text{gen}}$, $M_{ij}^{\text{imp}}$, and $M_{ij}^{\text{diff}}$ for same-subject same-key, different-subject same-key, and same-subject different-key pairs, respectively. The loss is as follows:

{\small
\begin{multline}
\mathcal{L}_{\text{disc}} = \frac{\sum_{i,j} M_{ij}^{\text{gen}} (S_{ij}^{\text{orig}} - S_{ij}^{\text{prot}})^2}{\sum M_{ij}^{\text{gen}}} 
+ \lambda_{\text{imp}} \cdot \frac{\sum_{i,j} M_{ij}^{\text{imp}} (S_{ij}^{\text{prot}})^2}{\sum M_{ij}^{\text{imp}}} + \\ \lambda_{\text{diff}} \cdot \frac{\sum_{i,j} M_{ij}^{\text{diff}} (S_{ij}^{\text{prot}})^2}{\sum M_{ij}^{\text{diff}}}
\end{multline}
}

coefficients $\lambda_{\text{imp}}$ and $\lambda_{\text{diff}}$ balance impostor separation and unlinkability.

\paragraph{Contrastive Loss: } This loss enhances class separability by pulling same-subject same-key pairs closer and pushing other pairs apart, using a margin $m$:

{\footnotesize
\begin{multline}
\mathcal{L}_{\text{contr}} = \frac{\sum_{i,j} M_{ij}^{\text{gen}} (1 - S_{ij}^{\text{prot}})}{\sum M_{ij}^{\text{gen}}} 
+ \beta_{\text{imp}} \cdot \frac{\sum_{i,j} M_{ij}^{\text{imp}} \max(S_{ij}^{\text{prot}} - m, 0)}{\sum M_{ij}^{\text{imp}}} \\ + \beta_{\text{other}} \cdot \frac{\sum_{i,j} (M_{ij}^{\text{diff}} + M_{ij}^{\text{mis}}) \max(S_{ij}^{\text{prot}} - m, 0)}{\sum (M_{ij}^{\text{diff}} + M_{ij}^{\text{mis}})}
\end{multline}
}

where $M_{ij}^{\text{mis}}$ denotes different-subject different-key pairs. Coefficients $\beta_{\text{imp}}$ and $\beta_{\text{other}}$ prioritize impostor pair separation.

\paragraph{Unlinkability Loss:} This loss ensures templates of the same subject with different keys ($k_1 \neq k_2$) are dissimilar, promoting renewability:
\begin{equation}
\mathcal{L}_{\text{unlink}} = |\cos(Z_T^{k_1}, Z_T^{k_2})|
\end{equation}

\paragraph{Key Diversity Loss:} This auxiliary loss ensures that keys yield diverse representations for the same identity:

\begin{equation}
\mathcal{L}_{\text{diverse}} =
\begin{cases}
0, & \text{if } k_1 = k_2 \\
\left| \cos \left( Z_T^{k_1}, Z_T^{k_2} \right) \right|, & \text{otherwise}
\end{cases}
\end{equation}

\paragraph{Total Objective:} A weighted sum of all loss components:
\begin{equation}
\mathcal{L}_{\text{total}} = 
\mathcal{L}_{\text{disc}} +
\mathcal{L}_{\text{contr}} +
\lambda_u \, \mathcal{L}_{\text{unlink}} +
\lambda_d \, \mathcal{L}_{\text{diverse}}
\end{equation}
where \( \lambda_u \) and \( \lambda_d \) control the strength of unlinkability and key diversity regularization. This unified formulation ensures high recognition fidelity while rigorously enforcing biometric privacy constraints.

\subsection{Recognition Module}
After biometric template protection, recognition is performed in the transformed domain. Given a query template $Z_T^q$ and an enrolled template $Z_T^e$, similarity is computed using cosine similarity, and a decision threshold $\theta$ is applied. \\
$\text{Match} \quad \text{if} \quad S > \theta, \quad \text{otherwise} \quad \text{Non-Match}$. \\
This approach ensures robust and privacy-preserving 3D face mesh recognition while mitigating the risk of template inversion and linkage attacks.

\section{Experiments}

\subsection{Experimental Settings}
\label{subsec:settings}

We evaluated the proposed GFT-GCN framework on two public 3D face datasets: BU-3DFE \cite{yin20063d} and FaceScape \cite{yang2020facescape}. BU-3DFE comprises 100 subjects with 25 expression scans each (2,500 meshes) under controlled conditions, whereas FaceScape includes 938 subjects with approximately 20 scans each (over 18,000 meshes), providing broader variability in pose, illumination, and morphology.  

For each dataset, subjects were divided into 70\% for training, 15\% for validation, and 15\% for testing. Details of dataset construction and pair generation are provided in Appendix~B. Input features were extracted from 3D face meshes using $n=10$ geometric descriptors, including 3D coordinates, vertex normals, dihedral angles, Gaussian curvature, and represented in a low-frequency spectral form $F_{low} \in \mathbb{R}^{k \times n}$, with $k \in \{10,20,25\}$. A spectral diffusion process was applied over $T$ steps ($T \in \{25,50,75\}$) to obtain privacy-preserving templates. The rationale for selecting $k$ and $T$ is discussed in Appendix~C. This setup ensures reproducibility while enabling a systematic study of recognition accuracy and privacy trade-offs across datasets.  

\subsection{Accuracy Performance Evaluation}
\begin{figure*}[ht]
     \center
     \includegraphics[scale=0.175]{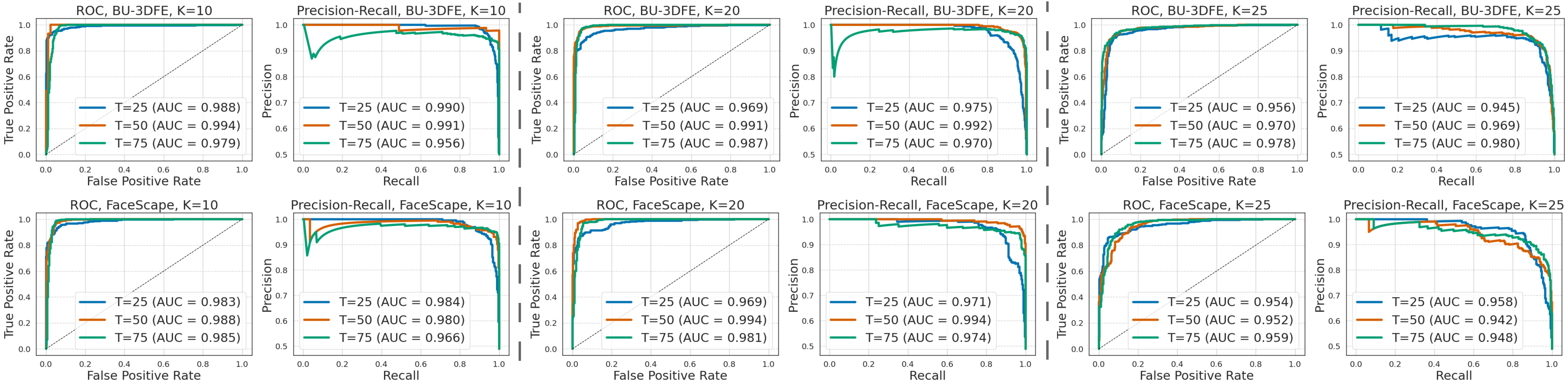}
     \caption{ROC and Precision-Recall curves for the GFT-GCN framework, confirming strong recognition performance.}
     \label{fig:roc}
\end{figure*}

The GFT-GCN framework achieves high recognition accuracy through a combination of advanced loss functions and feature preservation techniques. The Siamese contrastive loss applied to the GCN encourages embeddings of the same subject to be closely aligned while pushing embeddings of different subjects apart, thereby maximizing inter-subject separation and enhancing overall accuracy, Figure~\ref{fig:roc}. Additionally, the loss functions designed for spectral diffusion training ensure that the diffusion process maintains discriminative properties. The Graph Fourier Transform (GFT) plays a critical role by preserving key spectral features that distinguish individual identities.

\paragraph{Experiment 1: Baseline Accuracy vs. Protected Accuracy.}
We demonstrate that recognition accuracy remains high even after applying protection. Table~\ref{tab:evaluating_accuracy} presents the accuracy metrics, including Equal Error Rate (EER) and F1-score, for the GFT-GCN framework on the BU-3DFE and FaceScape datasets before and after protection across varying diffusion steps \( T \in \{25, 50, 75\} \) and spectral dimensions  \( K \in \{10, 20, 25\} \). The results indicate that the protection process does not significantly degrade the high baseline accuracy, confirming that the spectral diffusion process maintains recognition fidelity while enforcing privacy.

\begin{figure}[ht]
     \center
     \includegraphics[scale=0.18]{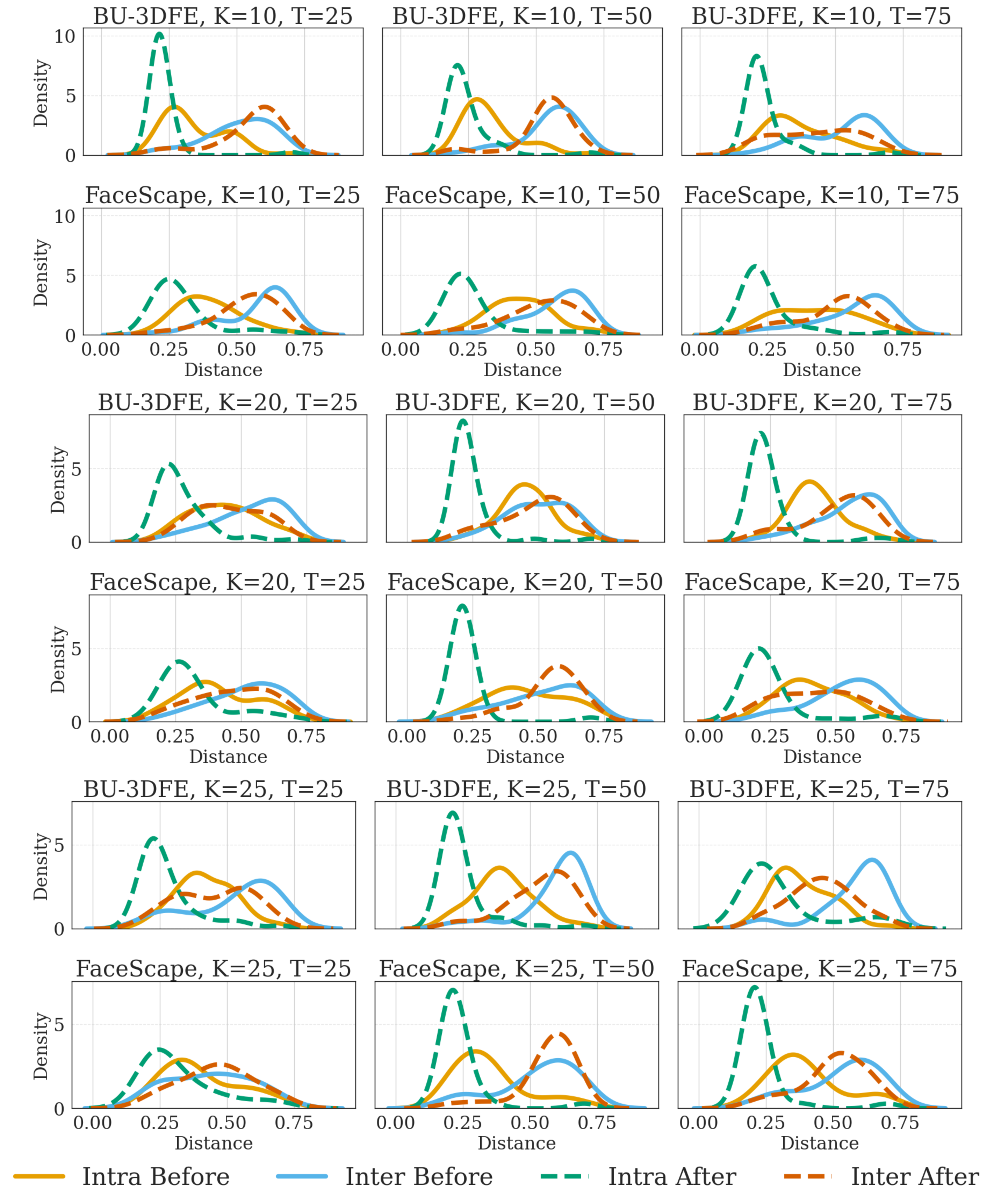}
     \caption{Intra-class and inter-class distance distributions before and after protection. Protected features remain well-separated across all settings, confirming that discriminability is preserved after diffusion.}
     \label{fig:intra_er_class}
\end{figure}

\begin{figure}[ht]
     \center
     \includegraphics[scale=0.27]{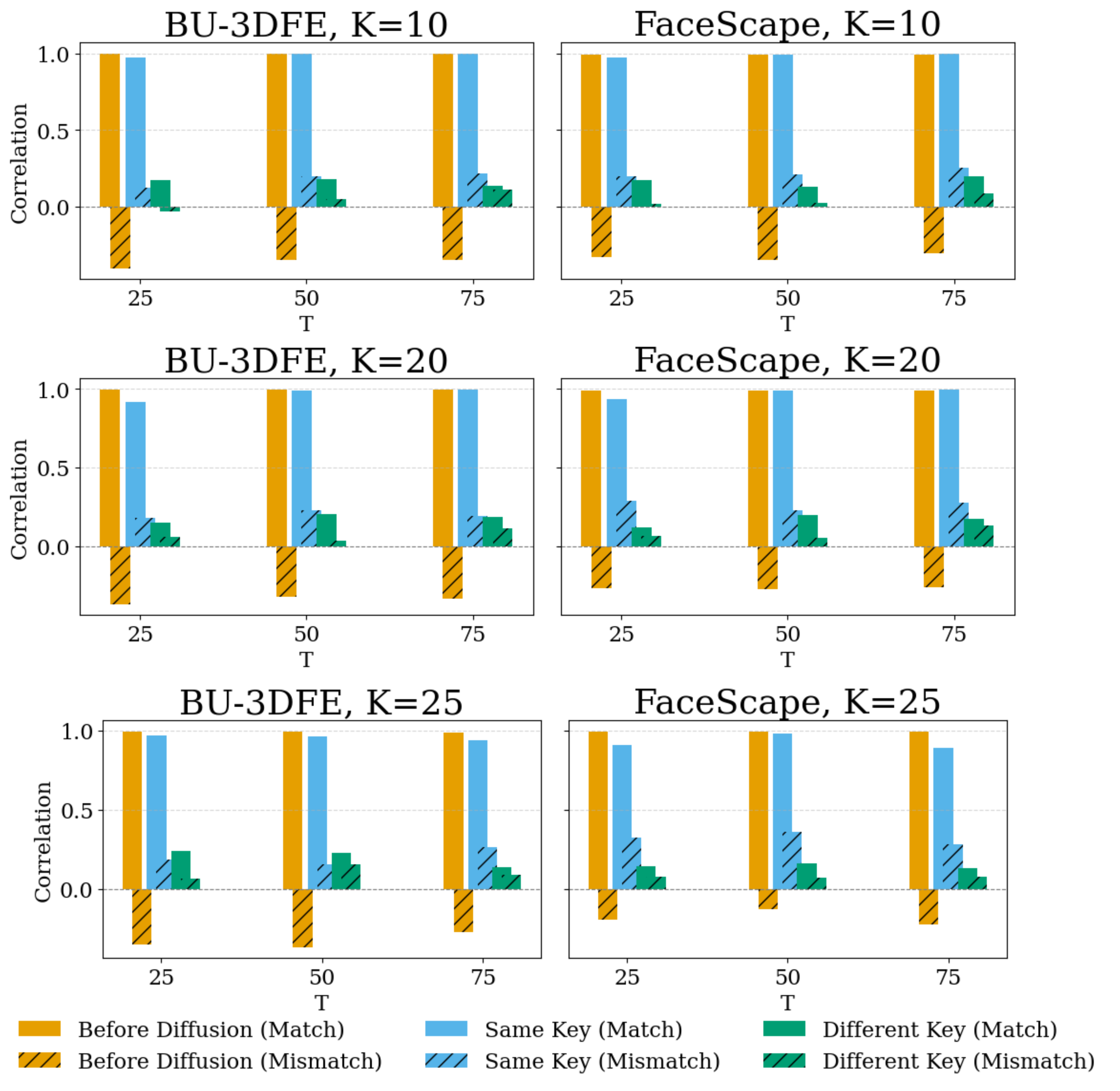}
     \caption{Correlation between biometric templates before and after diffusion under different key conditions. Protected templates retain high correlation for match pairs with the same key, while correlations drop significantly for mismatched and different key cases, confirming both renewability and unlinkability.}
     \label{fig:correlation}
\end{figure}

\paragraph{Experiment 2: Unprotected vs. Protected Feature Separability.}
We further validate that protected features retain their discriminative power by analyzing intra-class and inter-class distances before and after diffusion. Figure~\ref{fig:intra_er_class} illustrates the distance distributions, showing that protected features remain well-separated, supporting sustained high accuracy. Furthermore, the diffusion process improves intra-class compactness and inter-class separation, as evidenced by the maintained or enhanced distance metrics. Figure~\ref{fig:correlation} complements this with a correlation analysis of biometric templates before and after diffusion under different key conditions, reinforcing that the framework preserves separability and enhances privacy without compromising performance.

\begin{table*}[ht]
\centering 
\scriptsize

\caption{Evaluating the accuracy of the GFT-GCN framework before and after protection. For $K \in [10, 20, 25]$ and $T \in [25, 50, 75]$, the metrics (EER, $\theta$: best threshold, F1) are reported. The protection process preserves high accuracy with minimal degradation. Notably, the best performance is consistently achieved when $K = 10$ and $T = 50$, balancing discriminability and privacy.}
\label{tab:evaluating_accuracy}

\resizebox{0.9\textwidth}{!}{%
\begin{tabular}{@{}lcccc|ccccccccc@{}}
\toprule
 &
   &
  \multicolumn{3}{c|}{Before Protection} &
  \multicolumn{9}{c}{After Protection} \\ \midrule
\multicolumn{2}{l|}{T} &
  \multicolumn{3}{c|}{-} &
  \multicolumn{3}{c|}{25} &
  \multicolumn{3}{c|}{50} &
  \multicolumn{3}{c}{75} \\ \midrule
\multicolumn{1}{l|}{Dataset} &
  \multicolumn{1}{l|}{K} &
  EER$\downarrow$ &
  $\theta$ &
  F1$\uparrow$ &
  EER$\downarrow$ &
  $\theta$ &
  \multicolumn{1}{l|}{F1$\uparrow$} &
  ERR$\downarrow$ &
  $\theta$ &
  \multicolumn{1}{l|}{F1$\uparrow$} &
  EER$\downarrow$ &
  $\theta$ &
  F1 $\uparrow$ \\ \midrule
\multicolumn{1}{l|}{\multirow{3}{*}{BU-3DFE}} &
  \multicolumn{1}{l|}{10} &
  0.0000 &
  0.94 &
  1.000 &
  0.0467 &
  0.89 &
  \multicolumn{1}{l|}{0.9532} &
  \textbf{0.0144} &
  0.99 &
  \multicolumn{1}{l|}{\textbf{0.9846}} &
  0.0600 &
  0.96 &
  0.9603 \\
\multicolumn{1}{l|}{} &
  \multicolumn{1}{l|}{20} &
  0.0000 &
  0.82 &
  1.000 &
  0.0744 &
  0.83 &
  \multicolumn{1}{l|}{0.9256} &
  \textbf{0.0320} &
  0.95 &
  \multicolumn{1}{l|}{\textbf{0.9705}} &
  0.0708 &
  0.94 &
  0.9447 \\
\multicolumn{1}{l|}{} &
  \multicolumn{1}{l|}{25} &
  0.0000 &
  0.85 &
  1.000 &
  0.0844 &
  0.81 &
  \multicolumn{1}{l|}{0.9184} &
  \textbf{0.0722} &
  0.88 &
  \multicolumn{1}{l|}{\textbf{0.9317}} &
  0.0971 &
  0.79 &
  0.9061 \\ \midrule
\multicolumn{1}{l|}{\multirow{3}{*}{FaceScape}} &
  \multicolumn{1}{l|}{10} &
  0.0000 &
  0.84 &
  1.000 &
  0.0648 &
  0.81 &
  \multicolumn{1}{l|}{0.9386} &
  \textbf{0.0312} &
  0.97 &
  \multicolumn{1}{l|}{\textbf{0.9770}} &
  0.0426 &
  0.97 &
  0.9676 \\
\multicolumn{1}{l|}{} &
  \multicolumn{1}{l|}{20} &
  0.0000 &
  0.83 &
  1.000 &
  0.0837 &
  0.82 &
  \multicolumn{1}{l|}{0.9233} &
  \textbf{0.0423} &
  0.93 &
  \multicolumn{1}{l|}{\textbf{0.9690}} &
  0.0545 &
  0.98 &
  0.9553 \\
\multicolumn{1}{l|}{} &
  \multicolumn{1}{l|}{25} &
  0.0000 &
  0.89 &
  1.000 &
  0.0951 &
  0.78 &
  \multicolumn{1}{l|}{0.9105} &
  \textbf{0.0609} &
  0.88 &
  \multicolumn{1}{l|}{\textbf{0.9366}} &
  0.0722 &
  0.81 &
  0.9276 \\ \bottomrule
\end{tabular}%
}
\end{table*}

\subsection{Irreversibility Evaluation}
Irreversibility is a core security property of the GFT-GCN framework, ensured by our spectral diffusion process modeled as a Markov chain. Each state \( Z_t \) depends only on the previous state \( Z_{t-1} \) through the transition probability \( P(Z_t | Z_{t-1}) \). As the number of steps \( t \) increases, inversion becomes progressively more difficult. To recover the original template \( Z \), an attacker would need to solve:

\begin{equation}
    Z_{t-1} = Z_t - \sqrt{\beta_t}\epsilon - (1-\beta_t)\phi(Z_{t-1}, k)
\end{equation}

where \( \epsilon \) is unknown Gaussian noise drawn from a high-dimensional distribution and \( \phi \) is a trained neural network, making recovery intractable. With diffusion applied for \( T \) steps, the attacker must approximate:

\begin{equation}
    P(Z | Z_T) \approx \prod_{t=1}^T P(Z_{t-1} | Z_t)
\end{equation}

a multi-step inverse problem under uncertainty, which is NP-hard due to its combinatorial complexity.

\begin{figure}[ht]
     \center
     \includegraphics[scale=0.2]{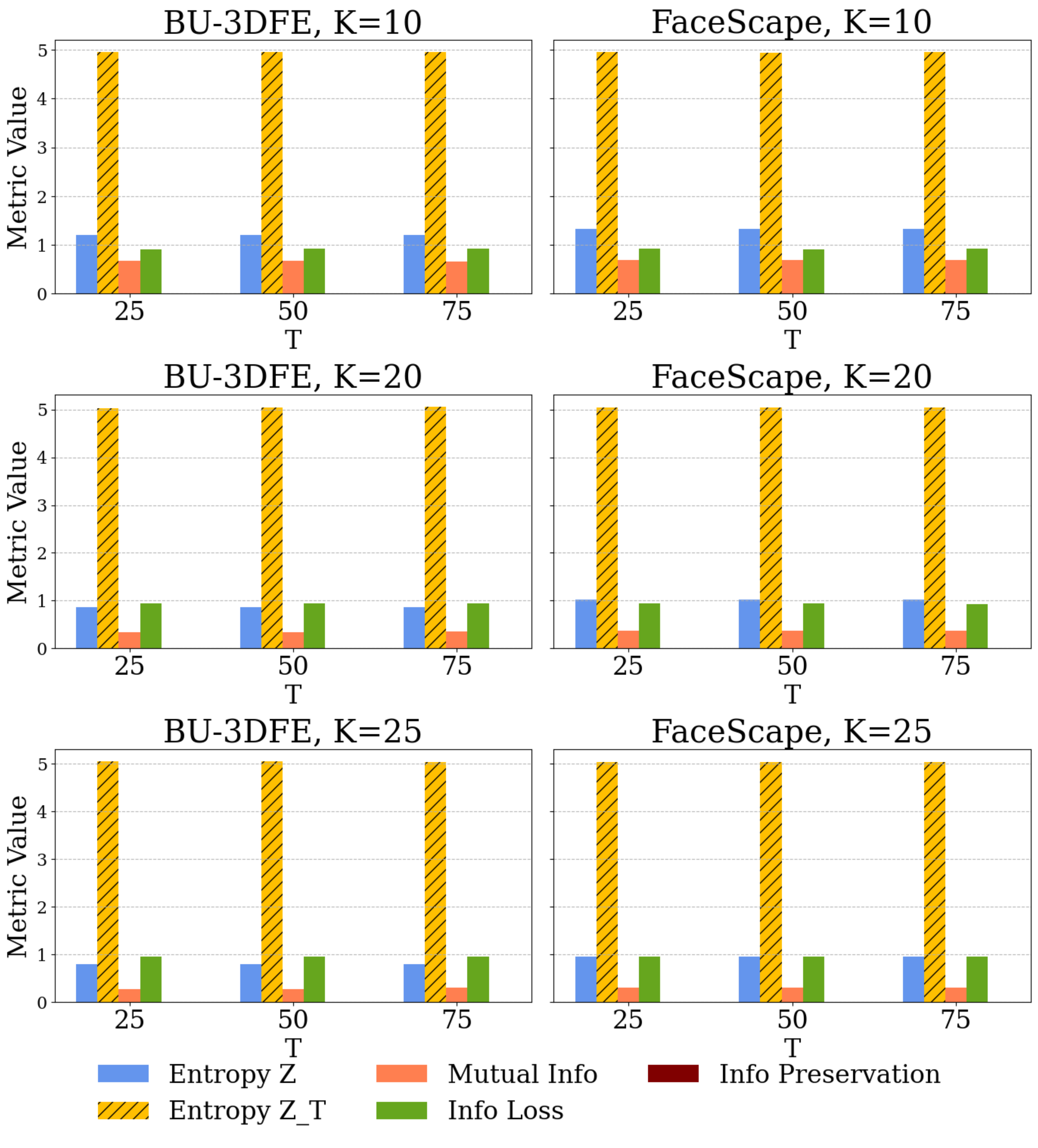}
     \caption{Comparison of the entropy \( H(Z) \) of unprotected templates, \( H(Z_T) \) of protected templates, mutual information, information loss, and information preservation. The significant increase in \( H(Z_T) \gg H(Z) \) highlights the loss of structured information, confirming the irreversibility of protected templates.}
     \label{fig:entropy}
\end{figure}

Our diffusion mechanism thus makes template reconstruction computationally infeasible. This is supported by entropy analysis (Figure~\ref{fig:entropy}), where the entropy of the protected template satisfies \( H(Z_T) \gg H(Z) \). The entropy growth reflects loss of invertible structure, making inversion impractical. Further, low mutual information (MI) between \( Z \) and \( Z_T \) minimizes leakage, while controlled preservation of discriminative features ensures reliable matching. Together, these results confirm that GFT-GCN provides strong resistance to inversion attacks while maintaining usability for recognition.

\subsection{Renewability and Unlinkability Evaluation}
The GFT-GCN framework ensures renewability and unlinkability, essential features for secure biometric applications. 

\textbf{Renewability:} From a single subject’s feature vector \( Z \), the framework can generate multiple distinct protected templates \( Z_{T1}, Z_{T2}, \ldots \) using different keys (\( key1, key2, \ldots \)). Each template remains valid for recognition. Since \( key1 \neq key2 \), the transformations satisfy \( \phi(Z, key1) \neq \phi(Z, key2) \), and the injected noise ensures \( Z_{T1} \neq Z_{T2} \). This confirms that unique, independent templates can always be issued for the same user (Figure~\ref{fig:correlation}).

\textbf{Unlinkability:} Templates generated from the same \( Z \) with different keys should remain statistically independent. We evaluate this by measuring

\begin{equation}
    Corr(Z_{T1}, Z_{T2}) = \frac{Cov(Z_{T1}, Z_{T2})}{\sqrt{Var(Z_{T1}) Var(Z_{T2})}}
\end{equation}

where: \\
\( Cov(Z_{T1}, Z_{T2}) = E[Z_{T1} Z_{T2}^T] - E[Z_{T1}]E[Z_{T2}^T] \approx 0 \). A near-zero covariance implies \( Corr(Z_{T1}, Z_{T2}) \approx 0 \), supporting independence. Figure~\ref{fig:correlation} shows that protected templates maintain high correlation for match pairs under the same key, but correlations drop sharply when keys differ or subjects mismatch. This validates that GFT-GCN achieves both renewability and unlinkability, ensuring strong privacy in 3D face recognition.

\subsection{Pre-image Attack Evaluation}
We evaluated the robustness of GFT-GCN against pre-image attacks \cite{dong2024security, nandakumar2015biometric, abdullahi2024biometric}, where an adversary seeks to reconstruct a plausible feature vector \( Z' \) from a protected template \( Z_T \). The goal is to find a pre-image whose protected form \( Z_T' = \text{template\_protection}(Z') \) maximizes similarity to the original \( Z_T \):  
\[
\arg\max_{Z'} \, \text{Similarity}(Z_T, Z_T')
\]

To simulate this scenario, we adopt the Constrained-Optimized Similarity-Based Attack (CSA) of Wang et al.~\cite{wang2021interpretable}. CSA performs iterative optimization to increase the similarity between \( Z_T \) and \( Z_T' \), while enforcing an \( \ell_2 \)-norm constraint on \( Z' \) to keep the reconstruction feasible.

We report performance using the Successful Attack Rate (SAR) at fixed thresholds (50 and 75) and at the best-performing threshold (\( \text{SAR}_{\text{Best}} \)). Results in Table~\ref{tab:preimage_attack} show that SAR values remain consistently low across all configurations, demonstrating that GFT-GCN effectively resists reconstruction-based attacks and provides strong protection against pre-image threats.

\begin{table}[ht]
\centering 
\scriptsize
\caption{Pre-image attack evaluation showing Successful Attack Rate (SAR) at thresholds 50, 75, and best ($SAR_{Best}$). Low SAR values confirm robustness against reconstruction attacks.}
\label{tab:preimage_attack}

\resizebox{0.95\columnwidth}{!}{%
\begin{tabular}{@{}llrrrlrrc@{}}
\toprule
\multicolumn{2}{l}{Metric}                  & \multicolumn{7}{c}{Dataset / K Value}                                   \\ \cmidrule(l){3-9} 
                    &                       & \multicolumn{3}{c}{BU-3DFE} &           & \multicolumn{3}{c}{FaceScape} \\ \cmidrule(lr){3-5} \cmidrule(l){7-9} 
\diagbox{T}{K} &
  \multicolumn{1}{c}{-} &
  \multicolumn{1}{c}{10} &
  \multicolumn{1}{c}{20} &
  \multicolumn{1}{c}{25} &
   &
  \multicolumn{1}{c}{10} &
  \multicolumn{1}{c}{20} &
  25 \\ \midrule
\multirow{3}{*}{25} & $SAR_{50}$            & 0.95    & 0.97    & 0.96    &           & 0.97     & 0.96     & 0.97    \\
                    & $SAR_{75}$            & 0.26    & 0.42    & 0.40    &           & 0.17     & 0.30     & 0.45    \\
                    & $SAR_{Best}$          & 0.00    & 0.05    & 0.02    & \textbf{} & 0.02     & 0.03     & 0.09    \\ \midrule
\multirow{3}{*}{50} & $SAR_{50}$            & 0.80    & 0.91    & 0.91    &           & 0.78     & 0.94     & 0.92    \\
                    & $SAR_{75}$            & 0.14    & 0.28    & 0.28    &           & 0.11     & 0.25     & 0.31    \\
                    & $SAR_{Best}$          & 0.00    & 0.00    & 0.01    & \textbf{} & 0.00     & 0.00     & 0.02    \\ \midrule
\multirow{3}{*}{75} & $SAR_{50}$            & 0.76    & 0.79    & 0.84    &           & 0.73     & 0.88     & 0.90    \\
                    & $SAR_{75}$            & 0.24    & 0.22    & 0.36    &           & 0.21     & 0.26     & 0.44    \\
                    & $SAR_{Best}$          & 0.00    & 0.00    & 0.11    & \textbf{} & 0.00     & 0.00     & 0.12    \\ \bottomrule
\end{tabular}%
}
\end{table}

\subsection{Ablation Study}
We conducted ablation experiments to identify the core components driving the performance of the GFT-GCN framework in 3D face recognition, focusing on the Graph Convolutional Network (GCN) and the Spectral Diffusion process.  

\textbf{Effect of GCN:} Removing the GCN leads to a marked drop in F1 scores both before and after protection (Table~\ref{tab:abl_study}), confirming its importance for extracting discriminative features from spectral inputs.  

\textbf{Effect of Spectral Diffusion:} Excluding Spectral Diffusion results in low template entropy \( H(Z) \) (Figure~\ref{fig:entropy}), reflecting weak transformation and poor protection. Spectral Diffusion also improves correlation patterns across different key conditions (Figure~\ref{fig:correlation}), which is critical for irreversibility and key-specific diversity.  

\textbf{Impact of Diffusion Steps:} Increasing the number of diffusion steps (e.g., \( T \geq 75 \)) further enhances privacy by injecting more noise, but this comes at the cost of reduced accuracy, highlighting the need to balance protection strength with recognition performance.  

These results demonstrate that both GCN and Spectral Diffusion are indispensable for achieving strong privacy guarantees without sacrificing recognition utility.

\begin{table}[ht]
\centering 
\scriptsize

\caption{Ablation study: F1 scores (\%) before and after protection when the GCN module is removed. These results underscore the importance of GCN in maintaining discriminative power, as performance degrades significantly when it is removed.}
\label{tab:abl_study}

\resizebox{\columnwidth}{!}{%
\begin{tabular}{@{}llc|ccc@{}}
\toprule
                      &                         & Before Protection & \multicolumn{3}{c}{After Protection} \\ \midrule
\multicolumn{1}{l|}{Dataset}                    & \multicolumn{1}{l|}{\diagbox{K}{T}} & -      & 25     & 50     & 75     \\ \midrule
\multicolumn{1}{l|}{\multirow{3}{*}{BU-3DFE}}   & \multicolumn{1}{l|}{10}                                & 0.6253 & 0.7209 & 0.7351 & 0.7295 \\
\multicolumn{1}{l|}{} & \multicolumn{1}{l|}{20} & 0.6322            & 0.7383     & 0.7644     & 0.7556     \\
\multicolumn{1}{l|}{} & \multicolumn{1}{l|}{25} & 0.6095            & 0.7292     & 0.7272     & 0.6977     \\ \midrule
\multicolumn{1}{l|}{\multirow{3}{*}{FaceScape}} & \multicolumn{1}{l|}{10}                                & 0.6467 & 0.7232 & 0.7302 & 0.7135 \\
\multicolumn{1}{l|}{} & \multicolumn{1}{l|}{20} & 0.6295            & 0.7294     & 0.7509     & 0.7392     \\
\multicolumn{1}{l|}{} & \multicolumn{1}{l|}{25} & 0.6097            & 0.7288     & 0.7464     & 0.7319     \\ \bottomrule
\end{tabular}%
}
\end{table}

\section{Conclusion and Future Directions}
We introduced GFT-GCN, a privacy-preserving framework for 3D face recognition that combines spectral feature extraction via the Graph Fourier Transform, geometric refinement using Graph Convolutional Networks, and a diffusion-based template protection mechanism. The framework produces irreversible, renewable, and unlinkable biometric templates while preserving high recognition accuracy. Experiments on BU-3DFE and FaceScape demonstrate that our method achieves a strong balance between security and utility, offering robust protection against reconstruction and pre-image attacks in realistic deployment scenarios.

For future work, we plan to optimize the diffusion process for faster computation, potentially using approximation methods, to enable real-time and large-scale applications. We also aim to integrate secure hardware and privacy-preserving techniques such as federated learning to strengthen client-side protection. Finally, incorporating formal guarantees through differential privacy could further reinforce security. These directions will extend the framework’s foundation, enhancing both efficiency and robustness for practical adoption.

\section{Acknowledgments}
This work was partially supported by JSPS KAKENHI Grants JP21H04907 and JP24H00732, by JST CREST Grants JPMJCR20D3 and JPMJCR2562 including AIP challenge program, by JST AIP Acceleration Grant JPMJCR24U3, and by JST K Program Grant JPMJKP24C2 Japan.

{
    \small
    \bibliographystyle{ieeenat_fullname}
    \bibliography{main}

@article{sardellitti2017graph,
  title={On the graph Fourier transform for directed graphs},
  author={Sardellitti, Stefania and Barbarossa, Sergio and Di Lorenzo, Paolo},
  journal={IEEE Journal of Selected Topics in Signal Processing},
  volume={11},
  number={6},
  pages={796--811},
  year={2017},
  publisher={IEEE}
}

@article{domingos2020graph,
  title={Graph Fourier transform: A stable approximation},
  author={Domingos, Joao and Moura, Jos{\'e} MF},
  journal={IEEE Transactions on Signal Processing},
  volume={68},
  pages={4422--4437},
  year={2020},
  publisher={IEEE}
}

@article{ricaud2019fourier,
  title={Fourier could be a data scientist: From graph Fourier transform to signal processing on graphs},
  author={Ricaud, Benjamin and Borgnat, Pierre and Tremblay, Nicolas and Gon{\c{c}}alves, Paulo and Vandergheynst, Pierre},
  journal={Comptes Rendus. Physique},
  volume={20},
  number={5},
  pages={474--488},
  year={2019}
}

@article{ho2020denoising,
  title={Denoising diffusion probabilistic models},
  author={Ho, Jonathan and Jain, Ajay and Abbeel, Pieter},
  journal={Advances in neural information processing systems},
  volume={33},
  pages={6840--6851},
  year={2020}
}

@inproceedings{nichol2021improved,
  title={Improved denoising diffusion probabilistic models},
  author={Nichol, Alexander Quinn and Dhariwal, Prafulla},
  booktitle={International conference on machine learning},
  pages={8162--8171},
  year={2021},
  organization={PMLR}
}

@inproceedings{friedrich2024wdm,
  title={Wdm: 3d wavelet diffusion models for high-resolution medical image synthesis},
  author={Friedrich, Paul and Wolleb, Julia and Bieder, Florentin and Durrer, Alicia and Cattin, Philippe C},
  booktitle={MICCAI Workshop on Deep Generative Models},
  pages={11--21},
  year={2024},
  organization={Springer}
}

@article{phillips2022spectral,
  title={Spectral diffusion processes},
  author={Phillips, Angus and Seror, Thomas and Hutchinson, Michael and De Bortoli, Valentin and Doucet, Arnaud and Mathieu, Emile},
  journal={arXiv preprint arXiv:2209.14125},
  year={2022}
}

@article{luo2023fast,
  title={Fast graph generation via spectral diffusion},
  author={Luo, Tianze and Mo, Zhanfeng and Pan, Sinno Jialin},
  journal={IEEE Transactions on Pattern Analysis and Machine Intelligence},
  volume={46},
  number={5},
  pages={3496--3508},
  year={2023},
  publisher={IEEE}
}

@inproceedings{yin20063d,
  title={A 3D facial expression database for facial behavior research},
  author={Yin, Lijun and Wei, Xiaozhou and Sun, Yi and Wang, Jun and Rosato, Matthew J},
  booktitle={7th international conference on automatic face and gesture recognition (FGR06)},
  pages={211--216},
  year={2006},
  organization={IEEE}
}

@inproceedings{yang2020facescape,
  title={Facescape: a large-scale high quality 3d face dataset and detailed riggable 3d face prediction},
  author={Yang, Haotian and Zhu, Hao and Wang, Yanru and Huang, Mingkai and Shen, Qiu and Yang, Ruigang and Cao, Xun},
  booktitle={Proceedings of the ieee/cvf conference on computer vision and pattern recognition},
  pages={601--610},
  year={2020}
}

@inproceedings{wang2021interpretable,
  title={Interpretable security analysis of cancellable biometrics using constrained-optimized similarity-based attack},
  author={Wang, Hanrui and Dong, Xingbo and Jin, Zhe and Teoh, Andrew Beng Jin and Tistarelli, Massimo},
  booktitle={Proceedings of the IEEE/CVF Winter Conference on Applications of Computer Vision},
  pages={70--77},
  year={2021}
}

@article{nandakumar2015biometric,
  title={Biometric template protection: Bridging the performance gap between theory and practice},
  author={Nandakumar, Karthik and Jain, Anil K},
  journal={IEEE Signal Processing Magazine},
  volume={32},
  number={5},
  pages={88--100},
  year={2015},
  publisher={IEEE}
}

@article{abdullahi2024biometric,
  title={Biometric template attacks and recent protection mechanisms: A survey},
  author={Abdullahi, Sani M and Sun, Shuifa and Wang, Beng and Wei, Ning and Wang, Hongxia},
  journal={Information Fusion},
  volume={103},
  pages={102144},
  year={2024},
  publisher={Elsevier}
}

@article{hahn2022biometric,
  title={Biometric template protection for neural-network-based face recognition systems: A survey of methods and evaluation techniques},
  author={Hahn, Vedrana Krivoku{\'c}a and Marcel, S{\'e}bastien},
  journal={IEEE Transactions on Information Forensics and Security},
  volume={18},
  pages={639--666},
  year={2022},
  publisher={IEEE}
}

@inproceedings{zhou2007template,
  title={Template protection and its implementation in 3D face recognition systems},
  author={Zhou, Xuebing},
  booktitle={Biometric technology for human identification IV},
  volume={6539},
  pages={192--199},
  year={2007},
  organization={SPIE}
}

@article{felouat20253ddgd,
  title={3DDGD: 3D Deepfake Generation and Detection Using 3D Face Meshes},
  author={Felouat, Hichem and Nguyen, Huy H and Yamagishi, Junichi and Echizen, Isao},
  journal={IEEE Access},
  volume={13},
  pages={107429--107441},
  year={2025},
  publisher={IEEE}
}

@inproceedings{papadopoulos2022face,
  title={Face-GCN: A graph convolutional network for 3D dynamic face recognition},
  author={Papadopoulos, Konstantinos and Kacem, Anis and Aouada, Djamila and others},
  booktitle={2022 8th International Conference on Virtual Reality (ICVR)},
  pages={454--458},
  year={2022},
  organization={IEEE}
}

@article{bernal2023review,
  title={A review on protection and cancelable techniques in biometric systems},
  author={Bernal-Romero, Juan Carlos and Ramirez-Cortes, Juan Manuel and Rangel-Magdaleno, Jose De Jesus and Gomez-Gil, Pilar and Peregrina-Barreto, Hayde and Cruz-Vega, Israel},
  journal={Ieee Access},
  volume={11},
  pages={8531--8568},
  year={2023},
  publisher={IEEE}
}

@article{kaur2023biometric,
  title={Biometric cryptosystems: a comprehensive survey},
  author={Kaur, Prabhjot and Kumar, Nitin and Singh, Maheep},
  journal={Multimedia Tools and Applications},
  volume={82},
  number={11},
  pages={16635--16690},
  year={2023},
  publisher={Springer}
}

@article{sharma2023survey,
  title={A survey on biometric cryptosystems and their applications},
  author={Sharma, Shreyansh and Saini, Anil and Chaudhury, Santanu},
  journal={Computers \& Security},
  volume={134},
  pages={103458},
  year={2023},
  publisher={Elsevier}
}

@article{osorio2021stable,
  title={Stable hash generation for efficient privacy-preserving face identification},
  author={Osorio-Roig, Dail{\'e} and Rathgeb, Christian and Drozdowski, Pawel and Busch, Christoph},
  journal={IEEE Transactions on Biometrics, Behavior, and Identity Science},
  volume={4},
  number={3},
  pages={333--348},
  year={2021},
  publisher={IEEE}
}

@article{guo20233d,
  title={3D face recognition: Two decades of progress and prospects},
  author={Guo, Yulan and Wang, Hanyun and Wang, Longguang and Lei, Yinjie and Liu, Li and Bennamoun, Mohammed},
  journal={ACM Computing Surveys},
  volume={56},
  number={3},
  pages={1--39},
  year={2023},
  publisher={ACM New York, NY}
}

@article{li2022comprehensive,
  title={A comprehensive survey on 3D face recognition methods},
  author={Li, Menghan and Huang, Bin and Tian, Guohui},
  journal={Engineering Applications of Artificial Intelligence},
  volume={110},
  pages={104669},
  year={2022},
  publisher={Elsevier}
}

@article{jing20233d,
  title={3D face recognition: A comprehensive survey in 2022},
  author={Jing, Yaping and Lu, Xuequan and Gao, Shang},
  journal={Computational Visual Media},
  volume={9},
  number={4},
  pages={657--685},
  year={2023},
  publisher={TUP}
}

@article{jing20213d,
  title={3D face recognition: A survey},
  author={Jing, Yaping and Lu, Xuequan and Gao, Shang},
  journal={arXiv preprint arXiv:2108.11082},
  year={2021}
}

@article{yang2024pointsurface,
  title={PointSurFace: Discriminative point cloud surface feature extraction for 3D face recognition},
  author={Yang, Junpeng and Li, Qiufu and Shen, Linlin},
  journal={Pattern Recognition},
  volume={156},
  pages={110858},
  year={2024},
  publisher={Elsevier}
}

@inproceedings{dommati2023real,
  title={Real-time 3D texture and motion analysis for face anti-spoofing using deep learning and computer vision},
  author={Dommati, Manikanta and Kiliroor, Cinu C},
  booktitle={International Conference on Recent Trends in Computing},
  pages={253--261},
  year={2023},
  organization={Springer}
}

@article{xu2024depth,
  title={Depth map denoising network and lightweight fusion network for enhanced 3D face recognition},
  author={Xu, Ruizhuo and Wang, Ke and Deng, Chao and Wang, Mei and Chen, Xi and Huang, Wenhui and Feng, Junlan and Deng, Weihong},
  journal={Pattern Recognition},
  volume={145},
  pages={109936},
  year={2024},
  publisher={Elsevier}
}

@article{yu2025adaptive,
  title={Adaptive representation learning and sample weighting for low-quality 3D face recognition},
  author={Yu, Cuican and Sun, Fengxun and Zhang, Zihui and Li, Huibin and Chen, Liming and Sun, Jian and Xu, Zongben},
  journal={Pattern Recognition},
  volume={159},
  pages={111161},
  year={2025},
  publisher={Elsevier}
}

@inproceedings{gilani2018learning,
  title={Learning from millions of 3D scans for large-scale 3D face recognition},
  author={Gilani, Syed Zulqarnain and Mian, Ajmal},
  booktitle={Proceedings of the IEEE conference on computer vision and pattern recognition},
  pages={1896--1905},
  year={2018}
}

@article{dong2024security,
  title={On the security risk of pre-image attack on cancelable biometrics},
  author={Dong, Xingbo and Park, Jaewoo and Jin, Zhe and Teoh, Andrew Beng Jin and Tistarelli, Massimo and Wong, Koksheik},
  journal={Journal of King Saud University-Computer and Information Sciences},
  volume={36},
  number={5},
  pages={102060},
  year={2024},
  publisher={Elsevier}
}

@article{dong2023laplacian2mesh,
  title={Laplacian2mesh: Laplacian-based mesh understanding},
  author={Dong, Qiujie and Wang, Zixiong and Li, Manyi and Gao, Junjie and Chen, Shuangmin and Shu, Zhenyu and Xin, Shiqing and Tu, Changhe and Wang, Wenping},
  journal={IEEE Transactions on Visualization and Computer Graphics},
  volume={30},
  number={7},
  pages={4349--4361},
  year={2023},
  publisher={IEEE}
}

@article{dong2024task,
  title={A task-driven network for mesh classification and semantic part segmentation},
  author={Dong, Qiujie and Gong, Xiaoran and Xu, Rui and Wang, Zixiong and Gao, Junjie and Chen, Shuangmin and Xin, Shiqing and Tu, Changhe and Wang, Wenping},
  journal={Computer Aided Geometric Design},
  volume={111},
  pages={102304},
  year={2024},
  publisher={Elsevier}
}

@inproceedings{kong2025not,
  title={Do Not DeepFake Me: Privacy-Preserving Neural 3D Head Reconstruction Without Sensitive Images},
  author={Kong, Jiayi and Song, Xurui and Huai, Shuo and Xu, Baixin and Luo, Jun and He, Ying},
  booktitle={Proceedings of the AAAI Conference on Artificial Intelligence},
  volume={39},
  number={4},
  pages={4383--4391},
  year={2025}
}

@inproceedings{yang2023towards,
  title={Towards effective adversarial textured 3d meshes on physical face recognition},
  author={Yang, Xiao and Liu, Chang and Xu, Longlong and Wang, Yikai and Dong, Yinpeng and Chen, Ning and Su, Hang and Zhu, Jun},
  booktitle={Proceedings of the IEEE/CVF conference on computer vision and pattern recognition},
  pages={4119--4128},
  year={2023}
}

@article{chang2024graph,
  title={Graph Fourier transform for spatial omics representation and analyses of complex organs},
  author={Chang, Yuzhou and Liu, Jixin and Jiang, Yi and Ma, Anjun and Yeo, Yao Yu and Guo, Qi and McNutt, Megan and Krull, Jordan E and Rodig, Scott J and Barouch, Dan H and others},
  journal={Nature Communications},
  volume={15},
  number={1},
  pages={7467},
  year={2024},
  publisher={Nature Publishing Group UK London}
}

@article{nakkina2024learnings,
  title={Learnings graph-Fourier spectra of textured surface images for defect localization},
  author={Nakkina, Tapan Ganatma and Karthikeyan, Adithyaa and Zhong, Yuhao and Eksin, Ceyhun and Bukkapatnam, Satish TS},
  journal={Manufacturing Letters},
  volume={41},
  pages={1568--1578},
  year={2024},
  publisher={Elsevier}
}

@article{ikeda2025speed,
  title={Speed-accuracy relations for diffusion models: Wisdom from nonequilibrium thermodynamics and optimal transport},
  author={Ikeda, Kotaro and Uda, Tomoya and Okanohara, Daisuke and Ito, Sosuke},
  journal={Physical Review X},
  volume={15},
  number={3},
  pages={031031},
  year={2025},
  publisher={APS}
}
}

\clearpage
\section*{Appendix}
\addcontentsline{toc}{section}{Appendix}
\appendix
\section{Diffusion Model}
This section clarifies the distinction between classical diffusion models and the proposed spectral variant. We emphasize that our framework does not use diffusion for generative purposes but for privacy-preserving feature transformation.  

\subsection{Classical Diffusion Models}
Classical diffusion models (e.g., Denoising Diffusion Probabilistic Models) iteratively corrupt data with Gaussian noise and learn a reverse process to recover the original input. Their primary goal is high-fidelity data generation and reconstruction of the input distribution.  

\subsection{Spectral Diffusion for Privacy}
In contrast, our spectral diffusion process is designed for biometric template protection, where reversibility is undesirable. We operate in the spectral domain of 3D face meshes, and the forward noising process is controlled through a trained neural transformation $\phi$. This process is supervised by discriminability and unlinkability losses, ensuring that privacy protection does not come at the cost of recognition accuracy.  

\subsection{Why Not Random Noise?}
Our approach differs fundamentally from arbitrary stochastic perturbations. Simply adding random noise may reduce accuracy and lead to irreversible degradation of biometric utility. Spectral diffusion, by contrast, provides:  
\begin{enumerate}
    \item \textbf{Controlled irreversibility:} Noise is injected stepwise in a Markovian process, 
    allowing predictable scaling of entropy with the number of steps $T$.  
    \item \textbf{Learned transformations:} The network $\phi$ shapes the trajectory of the noisy 
    features, making inversion computationally intractable while retaining discriminative 
    information.  
    \item \textbf{Balanced trade-off:} Loss functions enforce a balance between privacy 
    (irreversibility, unlinkability, renewability) and utility (high recognition accuracy).  
\end{enumerate}

The forward spectral diffusion process is not equivalent to arbitrary perturbation. It yields representations that are systematically hardened against reconstruction attacks while remaining useful for recognition tasks.

\section{Dataset Creation}
This section describes the creation and splitting of datasets used for training, validation, and testing in the GFT-GCN framework, ensuring a robust evaluation of the proposed 3D face mesh recognition method.  

\subsection{Dataset Composition}
For each dataset (e.g., BU-3DFE and FaceScape), let $N_{\text{subj}}$ denote the number of subjects and $N_{\text{fs}}$ the number of facial scans per subject. The total number of scans is given by
\[
N_{\text{total}} = N_{\text{subj}} \times N_{\text{fs}}
\]
 
\subsection{Example Generation}
For each client, we construct training examples as follows:
\begin{itemize}
    \item \textbf{Match pairs:} $N_{\text{fs}} \times (N_{\text{fs}} - 1)$ examples are generated by pairing scans from the same subject.  
    \item \textbf{Mismatch pairs:} An equal number of examples are created by pairing scans from one subject with scans from other subjects.  
\end{itemize}
This balanced strategy ensures that the dataset captures both intra-subject similarity and inter-subject differences, which is crucial for robust template protection.  

\subsection{Data Splits}
The dataset is divided into three subsets: 70\% for training, 15\% for validation, and 15\% for testing. The split is applied randomly across subjects to preserve diversity while preventing overlap between sets. This configuration provides sufficient data for optimization, enables reliable hyperparameter tuning, and ensures an unbiased evaluation of generalization performance. Importantly, the design reflects real-world variability in 3D facial data, supporting the framework’s privacy-preserving objectives.  

\begin{figure*}[ht]
     \center
     \includegraphics[scale=0.5]{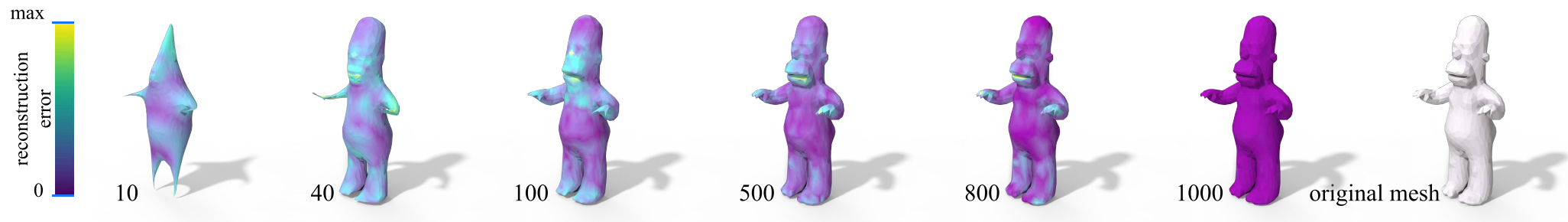}
     \caption{Effect of spectral dimensions \textit{K} on mesh reconstruction. Small \textit{K} captures only global shape, hiding identity details and improving privacy, while larger \textit{K} recovers local geometry, reducing privacy but improving fidelity; the original mesh is shown for reference ~\cite{dong2023laplacian2mesh}.}
     \label{fig:spectral_dim_K}
\end{figure*}

\section{Theoretical Analysis of Diffusion Steps and Spectral Dimensions}
This section provides a theoretical foundation for selecting the diffusion steps ($T = 25, 50, 75$) and spectral dimensions ($K = 10, 20, 25$) in the GFT-GCN framework. These parameters directly shape the \textbf{privacy--accuracy trade-off} in 3D face mesh recognition. The analysis draws on principles from diffusion models and Graph Fourier Transform (GFT) theory to justify the chosen ranges and quantify their impact on biometric template security and recognition performance.

\subsection{Diffusion Steps (\texorpdfstring{$T$}{T}) and Privacy}
Each diffusion step adds Gaussian noise (Eq.~4), increasing entropy $H(Z_T)$ in proportion to $T$, consistent with the Markovian formulation of diffusion models~\cite{ho2020denoising} ~\cite{ikeda2025speed}. As entropy grows, the inverse problem as transition probabilities $P(Z_t|Z_{t-1})$ becomes increasingly intractable, aligning with NP-hard reconstruction complexity and improving protection against pre-image attacks (Section~4.3).  

\subsection{Diffusion Steps (\texorpdfstring{$T$}{T}) and Accuracy}
More steps improve irreversibility, but may also degrade recognition performance by oversmoothing discriminative features. The spectral diffusion loss (Eq.~8) mitigates this by minimizing discriminability loss $L_{\text{disc}}$ (Eq.~4) across $T$ steps. Empirically, $T=50$ achieves the best balance, yielding $\text{EER} = 0.01$--$0.07\%$ and $\text{F1} = 0.93$--$0.98$ on BU-3DFE and FaceScape. The linear noise schedule $\beta_t$ preserves sufficient features for reliable cosine similarity matching (Section~4.2).  

\subsection{Spectral Dimensions (\texorpdfstring{$K$}{K}) and Privacy}
Keeping only low-frequency coefficients reduces reconstructive detail. From graph signal processing theory~\cite{sardellitti2017graph}, the first $K$ eigenvalues capture global structure while discarding high-frequency identity cues, Figure \ref{fig:spectral_dim_K}~\cite{dong2023laplacian2mesh}. For 10k-vertex meshes, $K=10$ yields an information loss factor of $\sim 6.9$, improving privacy relative to $K=25$ ($\sim 6.0$) ~\cite{chang2024graph} ~\cite{nakkina2024learnings}. Low mutual information results confirm this effect (Section~4.3).

\subsection{Spectral Dimensions (\texorpdfstring{$K$}{K}) and Accuracy}
Smaller $K$ boosts privacy but may discard useful identity information. The GCN refines these $K$-dimensional features ($F_{low} \in \mathbb{R}^{K \times n}$) under a Siamese contrastive loss (Eq.~2), maintaining class separability. $K=10$--$20$ preserves $\text{F1} \geq 0.96$, while larger $K$ increases cost without clear benefits, consistent with spectral dimensionality reduction~\cite{ricaud2019fourier}.  

\subsection{Trade-off Model}
We model how diffusion steps ($T$) and spectral dimensions ($K$) jointly affect privacy and accuracy.

\paragraph{Privacy:} Each diffusion step adds Gaussian noise, so the entropy of the protected template grows as
\[
\Delta H \;\approx\; \tfrac{K}{2}\log\!\big(1+cT\big) \;+\; C\log\!\left(\tfrac{N}{K}\right),
\]
where the first term reflects accumulated noise and the second term reflects information loss from truncating to $K$ coefficients ($N=|V|$ vertices). Larger $T$ and smaller $K$ increase privacy.

\paragraph{Accuracy:} Recognition depends on the signal-to-noise ratio of retained features. A simple approximation is
\[
\Delta \text{EER} \;\approx\; A \exp\!\Big(-\alpha\, \tfrac{K}{1+cT}\Big),
\]
Theoretically, the results show that accuracy improves with a larger $K$ and decreases with a larger $T$.


\begin{figure}[ht]
     \center
     \includegraphics[scale=0.27]{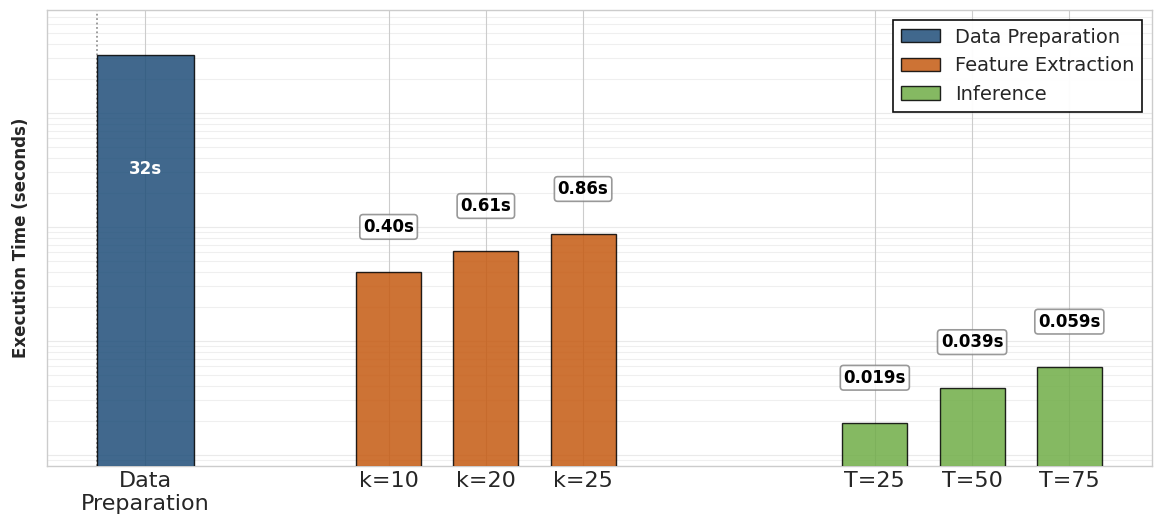}
     \caption{Execution time of GFT-GCN. Data preparation dominates (32s), feature extraction grows with spectral dimension (\(k\)), and inference remains lightweight across diffusion steps (\(T\)).}
     \label{fig:computational_costs}
\end{figure}

 \section{Computational Costs and Deployment}
We evaluated the efficiency of the proposed GFT-GCN framework in terms of computational cost and deployment feasibility. All experiments were conducted on NVIDIA Tesla A100 GPUs.  

\paragraph{Pipeline Stages:} The framework consists of three stages:  
\begin{enumerate}
    \item \textbf{Data Preparation:} Loading raw inputs, cropping the facial region, and normalizing the mesh.  
    \item \textbf{Feature Extraction:} Computing $n=10$ feature dimensions (3D coordinates, vertex normals, dihedral angles, and Gaussian curvature), and combining them with the low-frequency spectral form $F_{\text{low}}$.  
    \item \textbf{Inference:} Recognition using protected templates with diffusion-based template protection.  
\end{enumerate}

\paragraph{Cost Distribution:} The most expensive step is data preparation, which takes about 32s per example due to loading and preprocessing operations. The second contributor is feature extraction, dominated by eigenvalue and eigenvector computation of the graph Laplacian (0.40–0.86s depending on $K$). Inference is the lightest step, as diffusion adds only $T$ iterative operations (0.019–0.059s depending on $T$), Figure \ref{fig:computational_costs}.  

\paragraph{Deployment Feasibility:} With typical settings ($K=10$, $T=50$), the framework achieves near real-time performance, making it suitable for applications such as border control or secure device access. The client-server design further enhances scalability, since heavy computations (data preparation and feature extraction) are performed locally, reducing server overhead while maintaining privacy.  

\paragraph{Optimization Opportunities:} Additional efficiency gains may be achieved by accelerating eigenvalue computation through parallel solvers, or by reducing diffusion steps using adaptive noise schedules.

\end{document}